\documentclass[times,twocolumn,final,authoryear]{elsarticle}

\usepackage{amssymb}
\usepackage{latexsym}

\usepackage{url}
\usepackage{xcolor}
\definecolor{newcolor}{rgb}{.8,.349,.1}

\usepackage{amsmath}
\usepackage{pgfplots}
\pgfplotsset{width=\columnwidth,compat=1.8}
\usepackage{tikz}

\let\today\relax
\makeatletter
\def\ps@pprintTitle{%
    \let\@oddhead\@empty
    \let\@evenhead\@empty
    \def\@oddfoot{\footnotesize\itshape
         {Accepted in Pattern Recognition Letters} \hfill\today}%
    \let\@evenfoot\@oddfoot
    }
\makeatother

\journal{Pattern Recognition Letters}

\begin{document}

%\thispagestyle{empty}

%\ifpreprint
%  \vspace*{-1pc}
%\else
%  \vspace*{-6pc}
%\fi

%\begin{table*}[!t]
%\ifpreprint\else\vspace*{-15pc}%\fi

%\section*{Research Highlights (Required)}

%To create your highlights, please type the highlights against each
%\verb+\item+ command. 

%\vskip1pc

%\fboxsep=6pt
%\fbox{
%\begin{minipage}{.95\textwidth}
%It should be short collection of bullet points that convey the core
%findings of the article. It should  include 3 to 5 bullet points
%(maximum 85 characters, including spaces, per bullet point.)  
%\vskip1pc
%\begin{itemize}

% \item Diffusion-weighted images of the brain can provide meaningful information for biomarker investigation.

% \item These biomarkers can be used to support early diagnosis of cognitive impairment and to test novel treatments.

% \item Complex networks provide a powerful framework for studying the altered brain connectivity due to neurodegeneration.

% \item A novel method based on ensembling complex network `perspectives' with artificial neural networks is proposed.

% \item The effects of balancing issues on prediction accuracy are highlighted.

%\end{itemize}
%\vskip1pc
%\end{minipage}
%}

%\end{table*}

%\clearpage

\ifpreprint
  \setcounter{page}{1}
\else
  \setcounter{page}{1}
\fi

\begin{frontmatter}

\title{Ensembling complex network `perspectives' for mild cognitive impairment detection with artificial neural networks}

\author[1]{Eufemia Lella}
\author[2]{Gennaro Vessio\corref{cor1}
}
\cortext[cor1]{Corresponding author}
\ead{gennaro.vessio@uniba.it}

%\address[1]{Department of Physics, University of Bari, Bari, Italy}
%\address[2]{National Institute for Nuclear Physics, Bari, Italy}
\address[1]{Innovation Lab, Exprivia S.p.A., Molfetta, Italy}
\address[2]{Department of Computer Science, University of Bari, Bari, Italy}

%\received{}
%\finalform{}
%\accepted{}
%\availableonline{}
%\communicated{}

\begin{abstract}
In this paper, we propose a novel method for mild cognitive impairment detection based on jointly exploiting the complex network and the neural network paradigm. In particular, the method is based on ensembling different brain structural “perspectives” with artificial neural networks. On one hand, these perspectives are obtained with complex network measures tailored to describe the altered brain connectivity. In turn, the brain reconstruction is obtained by combining diffusion-weighted imaging (DWI) data to tractography algorithms. On the other hand, artificial neural networks provide a means to learn a mapping from topological properties of the brain to the presence or absence of cognitive decline. The effectiveness of the method is studied on a well-known benchmark data set in order to evaluate if it can provide an automatic tool to support the early disease diagnosis. Also, the effects of balancing issues are investigated to further assess the reliability of the complex network approach to DWI data.
\end{abstract}


\end{frontmatter}

%\linenumbers

\section{Introduction}
\label{sec1}

Mild cognitive impairment (MCI) is a non-disabling disorder characterized by an early state of abnormal cognitive function \citep{petersen2016mild}. An individual with MCI shows measurable changes in thinking skills---usually noticed by family members and friends---, which however do not affect the ability to carry out daily life activities. Nevertheless, 
according to some studies reported in literature, e.g. \citep{ward2013rate,vos2015prevalence}, people with MCI are at a higher risk of developing Alzheimer's disease (AD), or other kinds of dementia, than people without MCI. Research is ongoing to identify and validate useful \emph{biomarkers} that might indicate the risk of decline \citep{reitz2014alzheimer}. These may be used to support the early disease diagnosis and testing of novel treatments. 
Unfortunately, despite the large number of promising results, biological markers of MCI are at various stages of development and their use within standard clinical routines has not yet been established. 

Advances in this research have been obtained, in the last few years, in neuroimaging, particularly with diffusion-weighted imaging (DWI). DWI measures the water diffusion along the white matter (WM) fibers, thus it can provide meaningful information regarding their integrity \citep{amlien2014diffusion}. This information can highlight WM micro-structural changes that are related to neurodegeneration. In addition, when combined with tractography algorithms, DWI enables the reconstruction of the WM fiber tracts, providing a characterization of the physical connections of the brain that can be subsequently investigated through a complex network-based approach, e.g. \citep{lo2010diffusion}. In fact, the human brain can be modeled as a network the nodes of which are the anatomical regions, while edges represent the fiber tracts connecting them.

A recently proposed approach to study the diagnostic potentials of complex network measures consists in feeding these measures into supervised machine learning algorithms to automatize the disease detection, e.g., \citep{ebadi2017ensemble,schouten2017individual,lella2018communicability}. Developing a computerized decision support tool is desirable as it can provide a complementary approach to the standard evaluations which is non-invasive and low-cost. However, % an effective classification strategy tailored to support the early disease diagnosis in real-world settings has not yet been found. The problem is that, although
while models with very high prediction accuracy in detecting late-stage AD have been developed so far, the binary discrimination healthy/MCI is still hard. MCI, in fact, seems to be characterized by very minimal variations which makes it difficult to find meaningful patterns for distinguishing this state from normal aging. %This issue is encountered not only in the neuroimaing community (e.g., \textbf{citare}).
Furthermore, several existing works on this problem used either private data, e.g. \citep{wee2011enriched,ebadi2017ensemble}, or benchmark data with a disproportion of the MCI group compared to the control group, e.g. \citep{nir2015diffusion,prasad2015brain}. In the first case, the experimentation is not reproducible. In the second case, since machine learning algorithms are known to prefer the majority class when data are unbalanced, estimating sensitivity and specificity values may be biased by this factor. 

The contribution of this paper is two-fold. On one hand, we propose a novel classification strategy for the binary discrimination healthy/MCI. Starting from the observation that different complex network measures may provide different ``perspectives'' of the same networks under study, i.e.~they carry on non-redundant information, we develop an \emph{ensemble} model based on artificial neural networks each trained on different complex network features. Ensemble models are typically based on different classifiers which learn to predict the target output based on the same input. In the proposed method, instead, the ensemble is based on copies of the same classifier fed with different inputs. %The ensemble's performance is matched against the results provided by the individual features if taken alone and by their fusion in a single high-dimensional feature vector. 
%The ensemble is likely to provide better results than those obtained by the individual features if taken alone. 
%In addition, we provide the results of a feature importance analysis, based on the learning algorithm itself. The aim was to identify the brain regions the connectivity of which seems to be more related to the cognitive decline due to MCI. 
%The analysis is carried out on unbalanced benchmark data. %from the publicly available ADNI data set. 
On the other hand, this paper contributes by investigating the effects on classification performance of balancing the two groups under investigation. % both on the classification performance and the clinical validation of the selected features. 
To this end, some well-known under-sampling approaches are employed. To our best knowledge, this is the first attempt to study the impact of balancing issues to evaluate the reliability of the complex network approach to DWI data.

\begin{figure*}
    \centering
    \includegraphics[width=.8\textwidth]{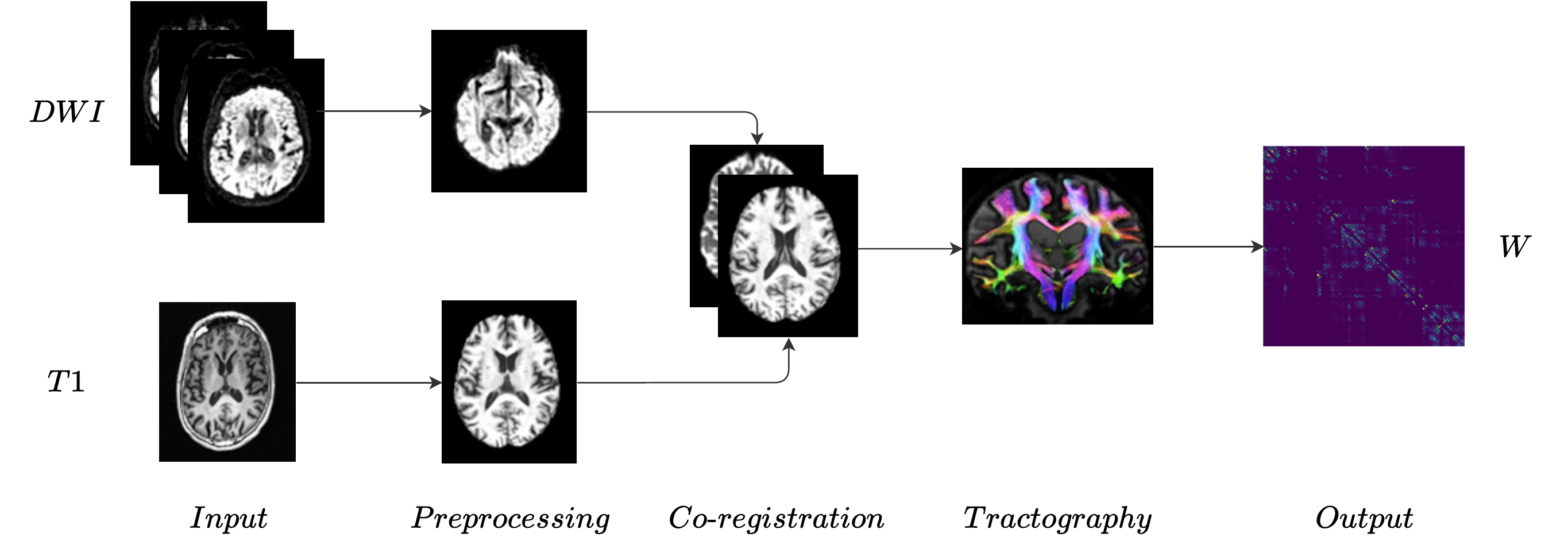}
    \caption{Image processing pipeline. DWI and T1 scans underwent several processing steps to obtain a weight connectivity matrix $W$ as final output \citep{lella2020communicability}. %More details in \citep{lella2019communicability}.
    }
    \label{fig:img_proc}
\end{figure*}

\section{Materials}

Data used in the preparation of this article were obtained from the Alzheimer's Disease Neuroimaging Initiative (ADNI) database (\url{adni.loni.usc.edu}). The ADNI was launched in 2003 as a public-private partnership, led by Principal Investigator Michael W. Weiner, MD. The primary goal of ADNI has been to test whether serial magnetic resonance imaging, positron emission tomography, other biological markers, and clinical and neuropsychological assessment can be combined to measure the progression of MCI and early AD. For up-to-date information, see \url{www.adni-info.org}.

In particular, we used an unbalanced cohort of 108 MCI patients and 49 healthy control (HC) subjects. Control subjects did not show signs of depression, MCI or dementia; MCI patients reported a subjective memory concern, but without any significant impairment in other cognitive domains: they preserved daily life activities with no sign of dementia. %It is worth to note that the MCI group is composed by \textbf{early/late MCI}. 
Scans were acquired using a 3-T GE Medical Systems scanner; more precisely, 46 separate images were acquired for each scan: 5 with negligible diffusion effects ($b_0$ images) and 41 diffusion-weighted images ($b=1000$ s/mm$^2$). For each subject, the T1 anatomical scan has also been used to perform tractography.

All procedures followed were in accordance with the ethical standards of the responsible committee on human experimentation (institutional and national) and with the Helsinki Declaration of 1975 and its later amendments. The ADNI project was approved by the Institutional Review Boards of all participating institutions and informed consent was obtained from all patients for being included in the study.

\section{Methods}

The analysis we carried out is described in the following subsections.

\subsection{Image Processing}

Image processing consisted in the reconstruction of the brain connectome from the DWI scans. For each subject, the DICOM images were acquired from ADNI and the dcm2nii tool, provided with the MRIcron suite (\url{https://www.nitrc.org/projects/mricron}), was used to convert them into the NIFTI format. The NIFTI images were then re-organized in the standard BIDS format. 

The subsequent processing steps, from image preprocessing to co-registration and structural connectome generation, were performed using tools provided by the FSL FMRIB Software Library (FSL) (\url{https://fsl.fmrib.ox.ac.uk/fsl/fslwiki}) and the MRtrix3 software package (\url{https://www.mrtrix.org/}). %For the sake of brevity, we omit here the detailed image processing pipeline the interested reader can found in \citep{lella2019communicability}. 
The main steps, which are well-established in the literature, are shown in Fig.~\ref{fig:img_proc}. First, a denoising step was performed to enhance the signal-to-noise ratio of the diffusion weighted signals so as to reduce the thermal noise. This noise is due to the stochastic thermal motion of the water molecules and their interaction with the surrounding micro-structure \citep{veraart2016denoising}. Head motion and eddy current distortions were corrected by aligning the DWI images of each subject to the average $b_0$ image. Then, the brain extraction tool (BET) was used for the skull-stripping of the brain \citep{smith2002fast}. The bias-field correction was used to correct all DWI volumes. Similarly, the T1 weighted scans were processed by performing the following steps: reorientation to the standard image MNI152, automatic cropping, bias-field correction, registration to the linear and nonlinear standard space and brain extraction. The next step was the inter-modal registration of the diffusion-weighted and T1 weighted images.

After preprocessing and co-registration, the structural connectome was generated. First, we generated a tissue-segmented image tailored to the anatomically constrained tractography \citep{zhang2001segmentation}. Then, we performed an unsupervised estimation of WM, gray matter and cerebro-spinal fluid. In the next step, the fiber orientation distributions for spherical deconvolution was estimated \citep{jeurissen2014multi}. We then performed a probabilistic tractography \citep{tournier2010improved} using dynamic seeding \citep{smith2015sift2} and anatomically-constrained tractography \citep{smith2012anatomically}, which improves the tractography reconstruction by using anatomical information through a dynamic thresholding strategy. We applied the spherical-deconvolution informed filtering of tractograms (SIFT2) methodology \citep{smith2015sift2}, which not only provides more biologically meaningful estimates of the structural connection density, but also a more efficient quantification of the streamlines connectivity. The obtained streamlines were mapped through a T1 parcellation scheme by using the AAL2 atlas \citep{rolls2015implementation}, which is a revised version of the automated anatomical atlas (AAL) including 120 regions. Finally, a robust structural connectome construction was performed for generating the connectivity matrices \citep{smith2015effects}. The pipeline here described has also been used in recent structural connectivity studies, for example \citep{amico2018mapping} and \citep{tipnis2018modeling}. 

The final output was a $120 \times 120$ weighted symmetric connectivity matrix $W$ for each subject, the entries $w_{ij}$ of which corresponded to the number of fiber tracts connecting the anatomical region $i$ to region $j$ in accordance with the AAL2 atlas.

\subsection*{Complex Network Features}

The connectivity matrix $W$ represents the structural complexity of the brain. From $W$ several graph measures can be computed to describe its topological properties. In this work, we consider three measures, namely the original weight information resulting from the application of tractography, shortest path length and weighted communicability. For each node pair $ij$, the shortest path length is simply the length of the shortest path from $i$ to $j$. It provides a different information from weights as it expresses the efficiency of the information flow. Analogously, communicability, firstly introduced by Estrada and Hatano \citep{estrada2008communicability}, then refined by Crofts and Higham \citep{crofts2009weighted} in the weighted case, is defined as:
\[
C_{ij} = \left( \text{exp} \left( D^{-1/2} W D^{-1/2} \right) \right)_{ij},
\]
where $D \in \mathbb{R}^{N \times N}$ is the diagonal strength matrix with $N$ the number of nodes. This network metric provides an even more general measure of the ease of communication inside the network, as it takes into account not only the shortest paths but all available routes connecting two nodes. 
Its usefulness in assessing the altered brain connectivity due to AD has been observed in recent works \citep{lella2018communicability,lella2019communicability}.

\begin{figure*}
    \centering
    \includegraphics[width=.65\textwidth]{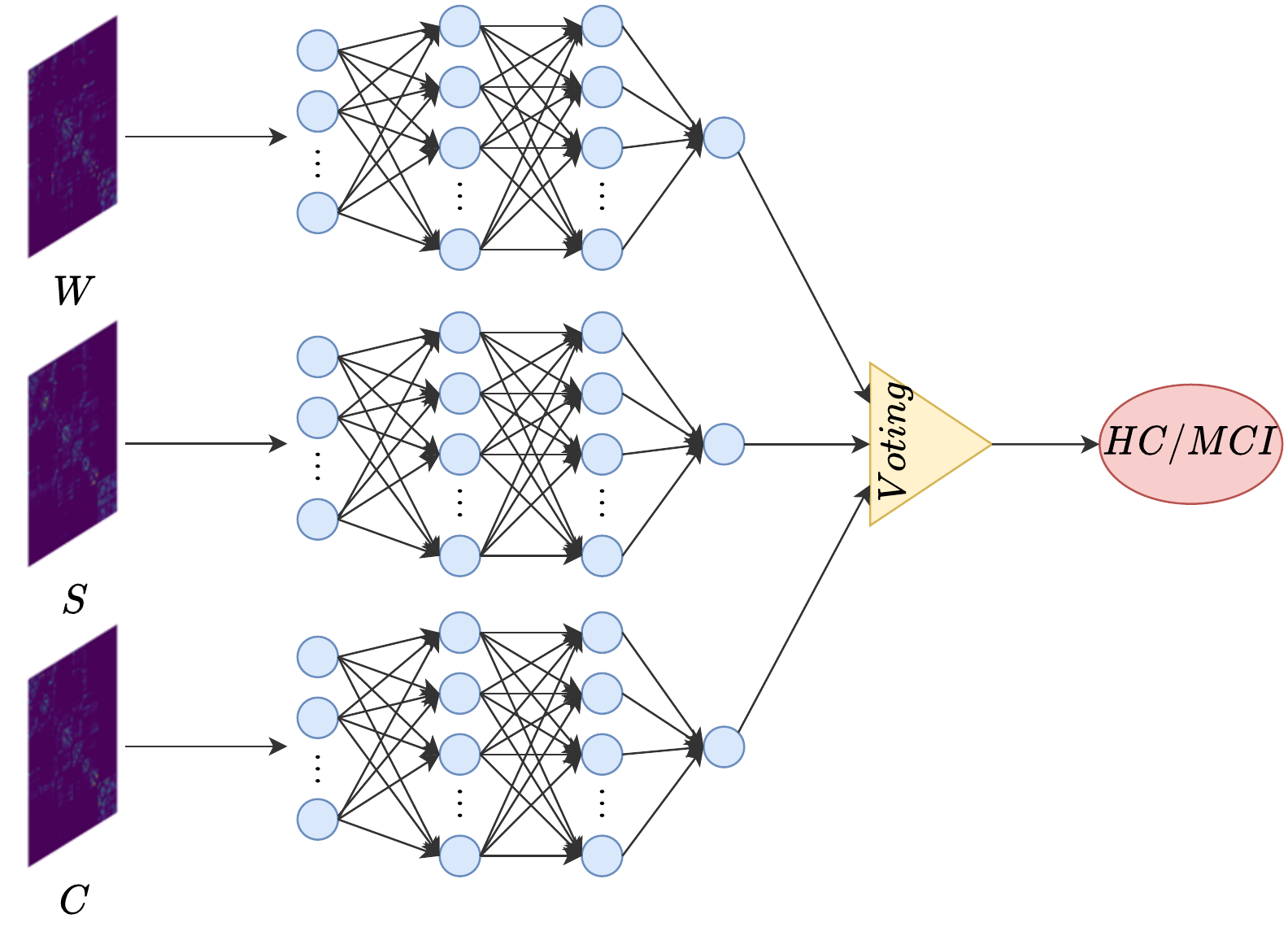}
    \caption{Proposed classification ensemble. The three complex network features are flattened and given as an input to three artificial neural networks sharing the same hyper-parameters. The final prediction is obtained as a soft voting of the individual classifiers' prediction. (Abbreviations: $W$ = weights; $S$ = shortest path length; $W$ = communicability; $HC$ = healthy controls; $MCI$ = mild cognitive impairment).}
    \label{fig:ensemble}
\end{figure*}

\subsection{Model Fitting}

%The classification strategies employed in the present research are depicted in Fig. \ref{fig:class_work}. The first strategy consisted in using as feature vector for feeding the classification model the individual 

Similar or conceptually different classification models can be combined through a voting scheme, so that the individual weaknesses of each classifier are mitigated \citep{theodoridis2015machine}. Different models trained independently, in fact, can look at slightly different aspects of the data, so having a chance to improve the predictions provided by each individual model in the ensemble. In particular, since different complex network measures are likely to provide different ``perspectives'' of the same networks under study, i.e.~they carry on non-redundant information, the ensemble of these features can take advantage of this diversification. In the present research, we developed an ensemble model based on the three different complex network features previously described: weights; shortest path length and communicability. As voting scheme, we used a \emph{soft} voting, which is based on averaging the probability scores given by the individual classifiers. Soft voting is usually preferred to \emph{hard} voting (which is based on a simple majority voting rule), as it takes into account more information than only the binary prediction, i.e.~the classifiers' uncertainty on the final decision they take. 

%An alternative approach consists in concatenating features into a single high-dimensional feature vector, one for each subject, which is then fed into a single classification algorithm. %However, such an approach could be detrimental to prediction accuracy, as it would result in a model more prone to overfitting. The ensemble over reduced feature vectors can mitigate this issue.

%To fairly evaluate the individual contribution of each single feature, as well as to fairly analyze how they support the detection of the mostly disease-related brain regions, we fixed the classification models used in the ensemble. 

As base learners, we used Multi-layer Perceptrons (MLPs) sharing the same hyper-parameters. Briefly speaking, an MLP is a feed-forward artificial neural network that can learn a nonlinear function approximator either for classification or regression \citep{murphy2018machine}. In contrast to traditional logistic regression, which is based on a single weighted linear combination between the input layer and the output layer, an MLP provides one or more nonlinear (\emph{hidden}) layers. In the present paper, we used an MLP with two hidden layers, 32 hidden units each; whereas, as activation function, we used the commonly used ReLU. Employing many more hidden layers would have had a negative impact on classification performance, given the disproportionately higher number of parameters to be optimized with respect to the number of training samples. Since the classification task is binary, the output layer performs a sigmoid activation:
\[
sigmoid(x) = \frac{1}{1+e^{-x}},
\]
where $x$ is the feature vector from the preceding hidden layer. The network attempts to minimize a classic cross-entropy loss function:
\[
\mathcal{H}(\theta) = \sum_{i=1}^{N} y_{i} \log (h_{\theta}(x_i)) + (1-y_i) \log (1-h_{\theta}(x_i)),
\]
where $\theta$ collectively indicates the parameters of the model, $N$ is the number of samples, and $y_i$ and $h_{\theta}(x_i)$ are the true and the predicted class label, respectively, for sample $x_i$. The network optimizes $\mathcal{H}(\theta)$ via backpropagation using the Limited-memory BFGS algorithm. This is an optimization algorithm in the family of quasi-Newton methods which is known to perform well when, as in our case, the data set is small \citep{morales2011remark}. It is worth noting that, since the behavior of the neural network can be heavily influenced by different feature scales, features were normalized in the range $\left[ 0, 1 \right]$ before training.

%Since our goal was to evaluate if the proposed approach provides improvements with respect to the predictions given by the individual features or by their fusion, for a fair comparison we provide experimental results obtained using the same classification model. 
We chose MLPs over other state-of-the-art classification algorithms mainly for two reasons. First, the use of nonlinearities within layers empowers the network with the capability of exploiting nonlinear relationships between data. Second, each layer can be equipped with a regularization term---the $\ell_2$ penalty in our case---which can help mitigate overfitting in presence of high-dimensional and possibly redundant features \citep{murphy2018machine}. The usefulness of neural network models in the clinical domain has been confirmed in several works, e.g. \citep{saniei2016parameter} and \citep{diaz2019dynamically}.  %Secondly, MLPs come with an embedded feature importance criterion which can be naturally employed for clinical-descriptive evaluations. More precisely, we used the well-known Gedeon method which computes a feature ranking by considering the weights connecting the input features to both the two hidden layers \citep{gedeon1997data}. These two characteristics make MLPs particularly useful for the classification problem at hand. 
An overall scheme of the proposed method is depicted in Fig.~\ref{fig:ensemble}.

\subsection{Balancing Strategies}

Most of machine learning methods are affected by the problem of having unbalanced data. %Balancing issues arise when the number of samples in each class are different. 
As the imbalance increases, the classification models tend to favor the correct prediction of the instances in the over-represented class. In a diagnostic problem, this may bias a reliable estimate of the system's accuracy, as a higher number of samples in the pathological (or control) group may result in an overoptimistic estimate of sensitivity (or specificity). This is an acknowledged issue in the machine learning community, e.g. \citep{haixiang2017learning}, and strategies to mitigate its effects for diagnostic purposes are sometimes adopted, e.g. \citep{duda2016use} and \citep{angelillo2019attentional}.

%In ADNI, which represents a benchmark data set, the MCI class is over-represented. 
In order to mitigate this issue, %and to measure how the imbalance affects the robustness of the method, 
we employed three under-sampling approaches:
\begin{itemize}
    \item \emph{Random under-sampling}: this is the most na\"{i}ve and easy way to under-sample the majority class. A subset sample of the over-represented class is selected randomly, then it is removed from the data set;
    \item \emph{Near miss-3}: this belongs to the ``near miss'' family \citep{mani2003knn} of methods which implements heuristics based on the $k$-nearest neighbors algorithm. The method proceeds in two steps: first, for each sample in the minority class, its $k$ nearest neighbors are detected; then, the examples in the majority class retained are the ones for which the average distance to the $k$ nearest neighbors is the largest. We set $k=3$. Near miss-3 is known to be less susceptible to noise than the other variants;
    \item \emph{Instance hardness threshold}: with this method, the MLP was trained and the subjects from the majority class for which the model returned the lower probability scores, i.e.~those for which the model was less confident on the label to be assigned, were removed \citep{smith2014instance}. 
\end{itemize}

While random under-sampling does not use any criterion to adjust the class distribution, near miss and instance hardness threshold focus on the data points that may be harder to classify. This can introduce a bias which may lead to overoptimistic results. %(\textbf{dire, nelle conclusioni, che non si tiene conto del rumore additivo (addendum: questo lo disse Angela e non ricordo che vuol dire)})

%Note that, for each re-sampling technique, we used the implementation provided in the \texttt{imbalanced-learn} Python toolbox \citep{JMLR:v18:16-365}. 

%\subsection{Feature Importance}

%\textbf{la feature importance e' un feature ranking, quindi non tiene conto delle interazioni fra le feature ma le valuta singolarmente: da verificare}

%\textbf{calcolarla ha senso perche', al contrario delle applicazioni delle ANN sulle immagini, qui le feature non sono raw pixel ma feature gia' ingegnerizzate}

%\textbf{l'importanza dei pesi attribuiti dal primo layer sarebbe disgregata dai layer successivi, ma qui, dopo il primo, ce n'e' solo un altro, quindi questo effetto e' poco pronunciato}

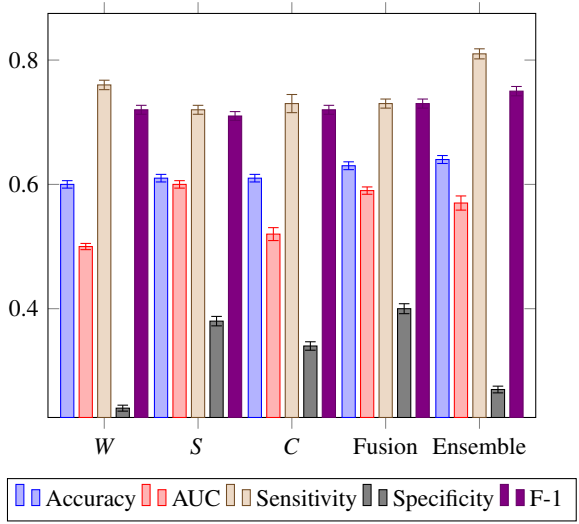
\begin{figure}[t]
\centering
\begin{tikzpicture}[font=\small]
\begin{axis}[
    ybar,
    bar width=5pt,
    ymin=0.3,
    ymax=0.8,
    enlargelimits=0.15,
    legend style={at={(0.5,-0.15)},anchor=north,legend columns=-1},
    symbolic x coords={$W$, $S$, $C$, Fusion, Ensemble},
    xtick=data,
    %nodes near coords,
    %nodes near coords align={vertical},
    ]
\addplot+ [error bars/.cd, y dir=both, y explicit relative] coordinates {($W$, 0.6) +- (0,0.01) ($S$, 0.61) +- (0,0.01) ($C$, 0.61) +- (0,0.01) (Fusion, 0.63) +- (0,0.01) (Ensemble, 0.64) +- (0,0.01)};
\addplot+ [error bars/.cd, y dir=both, y explicit relative] coordinates {($W$, 0.5) +- (0,0.01) ($S$, 0.6) +- (0,0.01) ($C$, 0.52) +- (0,0.02) (Fusion, 0.59) +- (0,0.01) (Ensemble, 0.57)+- (0,0.02)};
\addplot+ [error bars/.cd, y dir=both, y explicit relative] coordinates {($W$, 0.76) +- (0,0.01) ($S$, 0.72) +- (0,0.01) ($C$, 0.73) +- (0,0.02) (Fusion, 0.73) +- (0,0.01) (Ensemble, 0.81) +- (0,0.01)};
\addplot+ [error bars/.cd, y dir=both, y explicit relative] coordinates {($W$, 0.24) +- (0,0.02) ($S$, 0.38) +- (0,0.02) ($C$, 0.34) +- (0,0.02) (Fusion, 0.4) +- (0,0.02) (Ensemble, 0.27) +- (0,0.02)};
\addplot+ [error bars/.cd, y dir=both, y explicit relative] coordinates {($W$, 0.72) +- (0,0.01) ($S$, 0.71) +- (0,0.01) ($C$, 0.72) +- (0,0.01) (Fusion, 0.73) +- (0,0.01) (Ensemble, 0.75) +- (0,0.01)};
\legend{Accuracy, AUC, Sensitivity, Specificity, F-1}
\end{axis}
\end{tikzpicture}
\caption{Classification performance on the unbalanced data. (Abbreviations: $W$ = weights; $S$ = shortest path length; $W$ = communicability).}
\label{fig:unbalanced}
\end{figure}

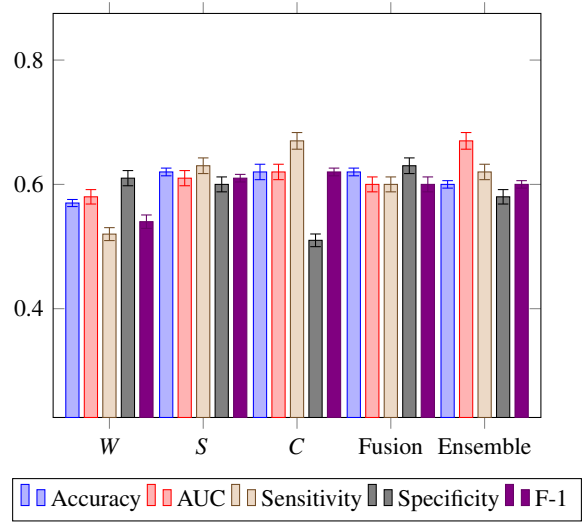
\begin{figure}[t]
\centering
\begin{tikzpicture}[font=\small]
\begin{axis}[
    ybar,
    bar width=5pt,
    ymin=0.3,
    ymax=0.8,
    enlargelimits=0.15,
    legend style={at={(0.5,-0.15)},anchor=north,legend columns=-1},
    symbolic x coords={$W$, $S$, $C$, Fusion, Ensemble},
    xtick=data,
    %nodes near coords,
    %nodes near coords align={vertical},
    ]
\addplot+ [error bars/.cd, y dir=both, y explicit relative] coordinates {($W$, 0.57) +- (0,0.01) ($S$, 0.62) +- (0,0.01) ($C$, 0.62) +- (0,0.02) (Fusion, 0.62) +- (0,0.01) (Ensemble, 0.60) +- (0,0.01)};
\addplot+ [error bars/.cd, y dir=both, y explicit relative] coordinates {($W$, 0.58) +- (0,0.02) ($S$, 0.61) +- (0,0.02) ($C$, 0.62) +- (0,0.02) (Fusion, 0.60) +- (0,0.02) (Ensemble, 0.67) +- (0,0.02)};
\addplot+ [error bars/.cd, y dir=both, y explicit relative] coordinates {($W$, 0.52) +- (0,0.02) ($S$, 0.63) +- (0,0.02) ($C$, 0.67) +- (0,0.02) (Fusion, 0.60) +- (0,0.02) (Ensemble, 0.62) +- (0,0.02)};
\addplot+ [error bars/.cd, y dir=both, y explicit relative] coordinates {($W$, 0.61) +- (0,0.02) ($S$, 0.60) +- (0,0.02) ($C$, 0.51) +- (0,0.02) (Fusion, 0.63) +- (0,0.02) (Ensemble, 0.58) +- (0,0.02)};
\addplot+ [error bars/.cd, y dir=both, y explicit relative] coordinates {($W$, 0.54) +- (0,0.02) ($S$, 0.61) +- (0,0.01) ($C$, 0.62) +- (0,0.01) (Fusion, 0.60) +- (0,0.02) (Ensemble, 0.60) +- (0,0.01)};
\legend{Accuracy, AUC, Sensitivity, Specificity, F-1}
\end{axis}
\end{tikzpicture}
\caption{Classification performance after random under-sampling. (Abbreviations: $W$ = weights; $S$ = shortest path length; $W$ = communicability).}
\label{fig:rus}
\end{figure}

\section{Experimental Results}

In this Section, we report the obtained results. 
As a baseline against which to compare the proposed method, we employed two approaches:
\begin{itemize}
    \item The first baseline consisted in evaluating the predictive accuracy of the individual features when separately fed to the neural network model;
    \item The second baseline was based on combining the three feature vectors into a single high-dimensional vector fed to the neural network. This strategy represents a typical complementary approach to that based on the ensemble.
\end{itemize}
For a fair comparison, the classification models used for the baseline experiments shared the same hyper-parameters of the MLP architecture used in the proposed ensemble.

%In the following, the results of two analysis are reported. The first one was aimed at evaluating the effectiveness of the proposed method in terms of classification performance. The second one was aimed at investigating the feature importance for a clinical validation. In both cases, the results obtained with the original data and after under-sampling are reported.

%It is worth noting that 
Since the set of data is small, we validated the classification performance through a %10 times repeated 
10-fold cross-validation. With this scheme, the set of data is partitioned into ten disjoint folds from which nine folds are used to train the learning algorithm, while the remaining fold is used to test it. This computation is iterated ten times, until each fold has been used as a test set once. 
In particular, we employed a \emph{stratified} cross-validation, so that each fold contained roughly the same number of subjects from each diagnostic group. The entire procedure was repeated ten times, with different permutations of the training and test samples, for a better generalization of the performance. 

As classification metrics, we used: %traditional performance metrics: 
accuracy; area under the ROC curve (AUC); sensitivity; specificity and F-1. We report the mean values of these metrics, averaged over all the cross-validation iterations. Also the standard errors are reported. In a clinical setting, sensitivity is one of the most important metrics to be monitored as it expresses the capability of the diagnostic tool to rule in disease when resulting in a positive response. 

In the following, both the results obtained with the original unbalanced data and those obtained after under-sampling are reported.

\subsection{Original Unbalanced Data}

Figure \ref{fig:unbalanced} shows the results obtained on the original unbalanced data (49 HC vs.~108 MCI). 
It can be observed that the performance of the individual features are quite comparable. As expected, the proposed ensemble generally improved upon the performance of the individual features, achieving a sensitivity of $0.81 \pm 0.01$ and an F-1 of $0.75 \pm 0.01$. The fusion of features, instead, provided little or no improvement over the single network measures. The sensitivity obtained with the ensemble was found to be statistically significant different from the other classification strategies (Mann-Whitney U test at the significance level $0.01$). The same applies to F-1, except for the comparison with the fusion of features ($p$-value $= 0.014$).

In general, very low values of specificity were obtained. As expected, this suggests that all classification models exhibited a preference for a more accurate prediction of the pathological group. This is confirmed by the lower values of AUC compared to sensitivity.  

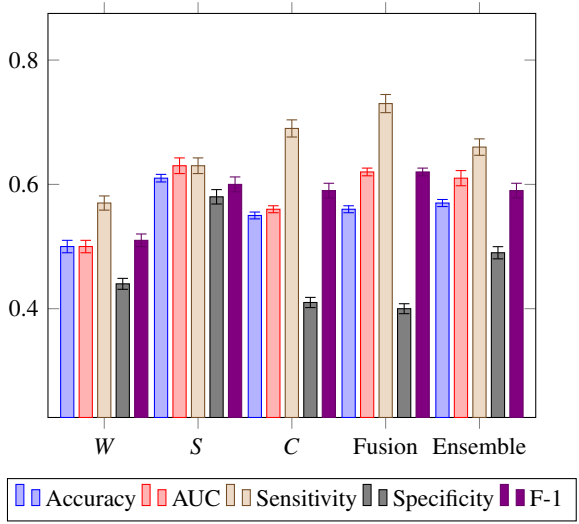
\begin{figure}[t]
\centering
\begin{tikzpicture}[font=\small]
\begin{axis}[
    ybar,
    bar width=5pt,
    ymin=0.3,
    ymax=0.8,
    enlargelimits=0.15,
    legend style={at={(0.5,-0.15)},anchor=north,legend columns=-1},
    symbolic x coords={$W$, $S$, $C$, Fusion, Ensemble},
    xtick=data,
    %nodes near coords,
    %nodes near coords align={vertical},
    ]
\addplot+ [error bars/.cd, y dir=both, y explicit relative] coordinates {($W$, 0.5) +- (0,0.02) ($S$, 0.61) +- (0,0.01) ($C$, 0.55) +- (0,0.01) (Fusion, 0.56) +- (0,0.01) (Ensemble, 0.57) +- (0,0.01)};
\addplot+ [error bars/.cd, y dir=both, y explicit relative] coordinates {($W$, 0.5) +- (0,0.02) ($S$, 0.63) +- (0,0.02) ($C$, 0.56) +- (0,0.01) (Fusion, 0.62) +- (0,0.01) (Ensemble, 0.61) +- (0,0.02)};
\addplot+ [error bars/.cd, y dir=both, y explicit relative] coordinates {($W$, 0.57) +- (0,0.02) ($S$, 0.63) +- (0,0.02) ($C$, 0.69) +- (0,0.02) (Fusion, 0.73) +- (0,0.02) (Ensemble, 0.66) +- (0,0.02)};
\addplot+ [error bars/.cd, y dir=both, y explicit relative] coordinates {($W$, 0.44) +- (0,0.02) ($S$, 0.58) +- (0,0.02) ($C$, 0.41) +- (0,0.02) (Fusion, 0.40) +- (0,0.02) (Ensemble, 0.49) +- (0,0.02)};
\addplot+ [error bars/.cd, y dir=both, y explicit relative] coordinates {($W$, 0.51) +- (0,0.02) ($S$, 0.6) +- (0,0.02) ($C$, 0.59) +- (0,0.02) (Fusion, 0.62) +- (0,0.01) (Ensemble, 0.59) +- (0,0.02)};
\legend{Accuracy, AUC, Sensitivity, Specificity, F-1}
\end{axis}
\end{tikzpicture}
\caption{Classification performance after near miss under-sampling. (Abbreviations: $W$ = weights; $S$ = shortest path length; $W$ = communicability).}
\label{fig:nm}
\end{figure}

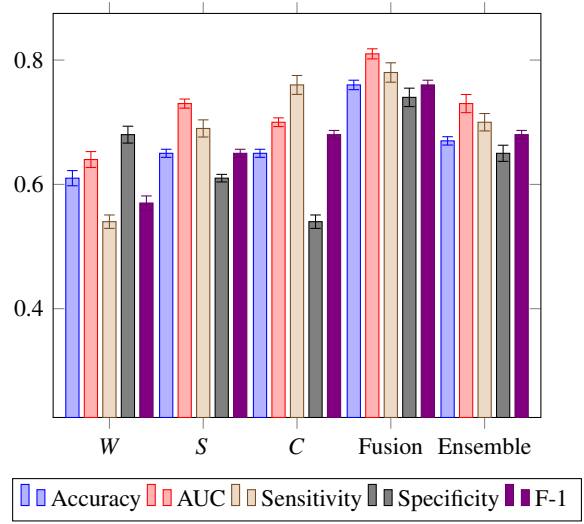
\begin{figure}[t]
\centering
\begin{tikzpicture}[font=\small]
\begin{axis}[
    ybar,
    bar width=5pt,
    ymin=0.3,
    ymax=0.8,
    enlargelimits=0.15,
    legend style={at={(0.5,-0.15)},anchor=north,legend columns=-1},
    symbolic x coords={$W$, $S$, $C$, Fusion, Ensemble},
    xtick=data,
    %nodes near coords,
    %nodes near coords align={vertical},
    ]
\addplot+ [error bars/.cd, y dir=both, y explicit relative] coordinates {($W$, 0.61) +- (0,0.02) ($S$, 0.65) +- (0,0.01) ($C$, 0.65) +- (0,0.01) (Fusion, 0.76) +- (0,0.01) (Ensemble, 0.67) +- (0,0.01)};
\addplot+ [error bars/.cd, y dir=both, y explicit relative] coordinates {($W$, 0.64) +- (0,0.02) ($S$, 0.73) +- (0,0.01) ($C$, 0.70) +- (0,0.01) (Fusion, 0.81) +- (0,0.01) (Ensemble, 0.73) +- (0,0.02)};
\addplot+ [error bars/.cd, y dir=both, y explicit relative] coordinates {($W$, 0.54) +- (0,0.02) ($S$, 0.69) +- (0,0.02) ($C$, 0.76) +- (0,0.02) (Fusion, 0.78) +- (0,0.02) (Ensemble, 0.70) +- (0,0.02)};
\addplot+ [error bars/.cd, y dir=both, y explicit relative] coordinates {($W$, 0.68) +- (0,0.02) ($S$, 0.61) +- (0,0.01) ($C$, 0.54) +- (0,0.02) (Fusion, 0.74) +- (0,0.02) (Ensemble, 0.65) +- (0,0.02)};
\addplot+ [error bars/.cd, y dir=both, y explicit relative] coordinates {($W$, 0.57) +- (0,0.02) ($S$, 0.65) +- (0,0.01) ($C$, 0.68) +- (0,0.01) (Fusion, 0.76) +- (0,0.01) (Ensemble, 0.68) +- (0,0.01)};
\legend{Accuracy, AUC, Sensitivity, Specificity, F-1}
\end{axis}
\end{tikzpicture}
\caption{Classification performance after instance hardness threshold under-sampling. (Abbreviations: $W$ = weights; $S$ = shortest path length; $W$ = communicability).}
\label{fig:iht}
\end{figure}

\subsection{Balanced Data}

In Fig.~\ref{fig:rus}, the results obtained after randomly re-sampling the majority class are shown. An overall performance decrease can be observed. The best results were obtained by communicability (sensitivity $= 0.67 \pm 0.02$) and the ensemble (AUC $= 0.67 \pm 0.02$). These values were statistically significant different from the other classification strategies (Mann-Whitney U test at the significance level $0.05$). Interestingly, shortest path length, communicability and the ensemble of features exhibited a sensitivity higher than specificity. This trend was reversed in the case of weights and the fusion of features. 

When near miss under-sampling was used, some metrics slightly improved with respect to random under-sampling. Once again, communicability provided the best sensitivity over the other individual features (i.e., a mean value of $0.69 \pm 0.02$). The ensemble and the fusion of features show similar accuracy and AUC; whereas, concerning the other metrics,  a significantly ($p$-value $= 0.009$) higher sensitivity was achieved by the fusion of features (i.e., a mean value of $0.73 \pm 0.02$).

Finally, instance hardness threshold under-sampling provided an overall improvement over the other re-sampling techniques. This was expected, as this technique removes the data points that are harder to classify. Among the individual features, shortest path length and communicability provided the best results, with communicability achieving a sensitivity of $0.76 \pm 0.02$. The overall best performance, instead, were obtained by the fusion of features (i.e., a mean AUC of $0.81 \pm 0.01$ and a mean sensitivity of $0.78 \pm 0.02$). These performance were statistically significant different from the other classification strategies (Mann-Whitney U test at the significance level 0.01), except for sensitivity in the comparison with communicability ($p$-value $= 0.102$) and specificity in the comparison with weights ($p$-value $= 0.032$).

For all the three balanced data sets, although with performance lower compared to the unbalanced data, all classification strategies were almost always better in detecting the pathological condition in the pathological group. This highlights the robustness of the complex network approach in combination with the neural network paradigm against the type II error.

\section{Conclusion}

In this work, a novel classification method for MCI detection based on DWI data 
has been proposed. The method is based on ensembling neural network models fed with different graph measures. These measures can provide non-overlapping information on the same graphs under study, so the ensemble can benefit from this diversification. The proposed method exhibited good sensitivity either when using unbalanced or balanced groups. In fact, in this paper we have also shown the detrimental effects on classification performance when the pathological and the control group are equally represented.

Future works should attempt to identify the brain regions the connectivity of which is more related to the cognitive decline due to MCI. To this end, a feature importance analysis, based on the learning algorithm itself, could be done.

\section*{Conflict of Interest}

The authors declare no conflict of interest.

\section*{Acknowledgments}

Data collection and sharing for this project was funded by the Alzheimer's Disease Neuroimaging Initiative (ADNI) (National Institutes of Health Grant U01 AG024904) and DOD ADNI (Department of Defense award number W81XWH-12-2-0012). ADNI is funded by the National Institute on Aging, the National Institute of Biomedical Imaging and Bioengineering, and through generous contributions from the following: AbbVie, Alzheimer's Association; Alzheimer's Drug Discovery Foundation; Araclon Biotech; BioClinica, Inc.; Biogen; Bristol-Myers Squibb Company; CereSpir, Inc.; Cogstate; Eisai Inc.; Elan Pharmaceuticals, Inc.; Eli Lilly and Company; EuroImmun; F. Hoffmann-La Roche Ltd and its affiliated company Genentech, Inc.; Fujirebio; GE Healthcare; IXICO Ltd.; Janssen Alzheimer Immunotherapy Research \& Development, LLC.; Johnson \& Johnson Pharmaceutical Research \& Development LLC.; Lumosity; Lundbeck; Merck \& Co., Inc.; Meso Scale Diagnostics, LLC.; NeuroRx Research; Neurotrack Technologies; Novartis Pharmaceuticals Corporation; Pfizer Inc.; Piramal Imaging; Servier; Takeda Pharmaceutical Company; and Transition Therapeutics. The Canadian Institutes of Health Research is providing funds to support ADNI clinical sites in Canada. Private sector contributions are facilitated by the Foundation for the National Institutes of Health (www.fnih.org). The grantee organization is the Northern California Institute for Research and Education, and the study is coordinated by the Alzheimer's Therapeutic Research Institute at the University of Southern California. ADNI data are disseminated by the Laboratory for Neuro Imaging at the University of Southern California.
%Image processing was partially carried out by using the facilities of the ReCaS data center (\url{http://www.recas-bari.it}).

\bibliographystyle{model2-names}
\bibliography{refs}

\end{document}